\documentclass[runningheads]{llncs}

\usepackage{eccv}

\usepackage{eccvabbrv}

\usepackage{graphicx}
\usepackage{booktabs}

\usepackage[accsupp]{axessibility}  

\usepackage[pagebackref,breaklinks,colorlinks,citecolor=eccvblue]{hyperref}

\usepackage{orcidlink}
\usepackage{shadowtext}
\usepackage{tabularx}
\usepackage{xcolor}
\usepackage{colortbl}
\usepackage{bm}
\usepackage{multirow}
\usepackage{multicol}
\usepackage{algorithm}
\usepackage{algpseudocode}
\usepackage{amsmath}

\usepackage{float}
\usepackage{marvosym}
\usepackage{dsfont}

\footnotetext[2]{\textsuperscript{\Letter} Corresponding author.}

\newcolumntype{C}[1]{>{\centering\arraybackslash}m{#1}}
\newcolumntype{Y}{>{\centering\arraybackslash}X}

\begin{document}

\title{AdaLog: Post-Training Quantization for Vision Transformers with Adaptive Logarithm Quantizer} 

\titlerunning{Adaptive Logarithm Quantizer}

\author{Zhuguanyu Wu\inst{1,2}\orcidlink{0009-0008-2183-9979} \and
Jiaxin Chen\inst{1,2}\orcidlink{0000-0002-0112-4166}\textsuperscript{(\Letter)} 
\and
Hanwen Zhong\inst{1,2}\orcidlink{0009-0003-9067-071X} \and
Di Huang\inst{2}\orcidlink{0000-0002-2412-9330} \and
Yunhong Wang\inst{1,2}\orcidlink{0000-0001-8001-2703}
}
\authorrunning{Z.~Wu et al.}

\institute{State Key Laboratory of Virtual Reality Technology and Systems, Beihang University, Beijing, China \and
School of Computer Science and Engineering, Beihang University, Beijing, China
\email{\{goatwu,jiaxinchen,hanwenzhong,dhuang,yhwang\}@buaa.edu.cn}}

\maketitle

\begin{abstract}
Vision Transformer (ViT) has become one of the most prevailing fundamental backbone networks in the computer vision community. Despite the high accuracy, deploying it in real applications raises critical challenges including the high computational cost and inference latency. Recently, the post-training quantization (PTQ) technique has emerged as a promising way to enhance ViT's efficiency. Nevertheless, existing PTQ approaches for ViT suffer from the inflexible quantization on the post-Softmax and post-GELU activations that obey the power-law-like distributions. To address these issues, we propose a novel non-uniform quantizer, dubbed the Adaptive Logarithm AdaLog (AdaLog) quantizer. It optimizes the logarithmic base to accommodate the power-law-like distribution of activations, while simultaneously allowing for hardware-friendly quantization and de-quantization. By employing the bias reparameterization, the AdaLog quantizer is applicable to both the post-Softmax and post-GELU activations. Moreover, we develop an efficient Fast Progressive Combining Search (FPCS) strategy to determine the optimal logarithm base for AdaLog, as well as the scaling factors and zero points for the uniform quantizers. Extensive experimental results on public benchmarks demonstrate the effectiveness of our approach for various ViT-based architectures and vision tasks including classification, object detection, and instance segmentation. Code is available at \url{https://github.com/GoatWu/AdaLog}.
\end{abstract}

\section{Introduction}
\label{sec:intro}

\begin{figure}[t]
    \centering
    \begin{subfigure}{.36\columnwidth}
        \centering
        \includegraphics[width=.98\linewidth]{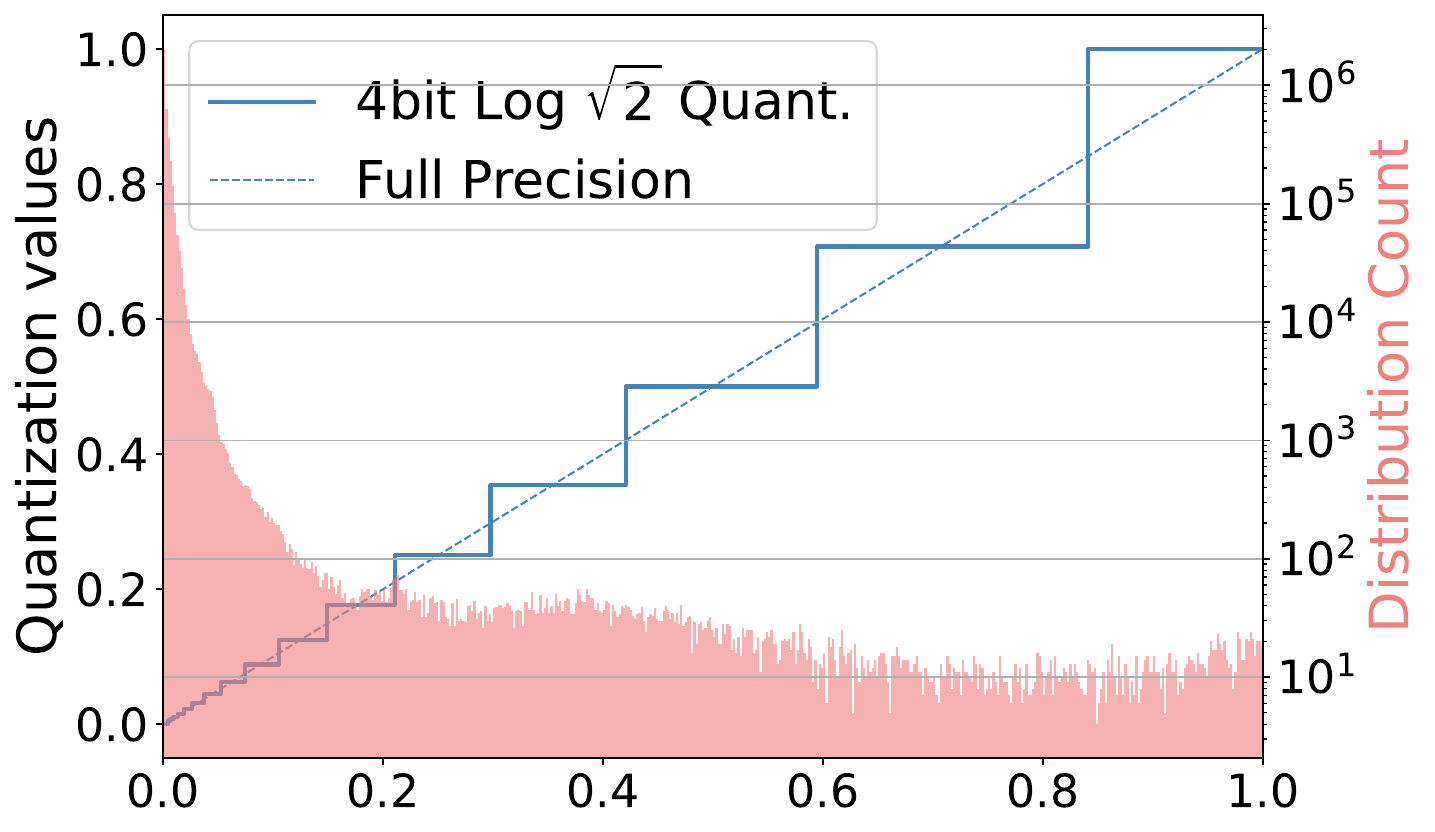}
        \caption{4-bit log$\sqrt{2}$ Quantizer}
        \label{fig:sub1}
    \end{subfigure}%
    \begin{subfigure}{.36\columnwidth}
        \centering
        \includegraphics[width=.98\linewidth]{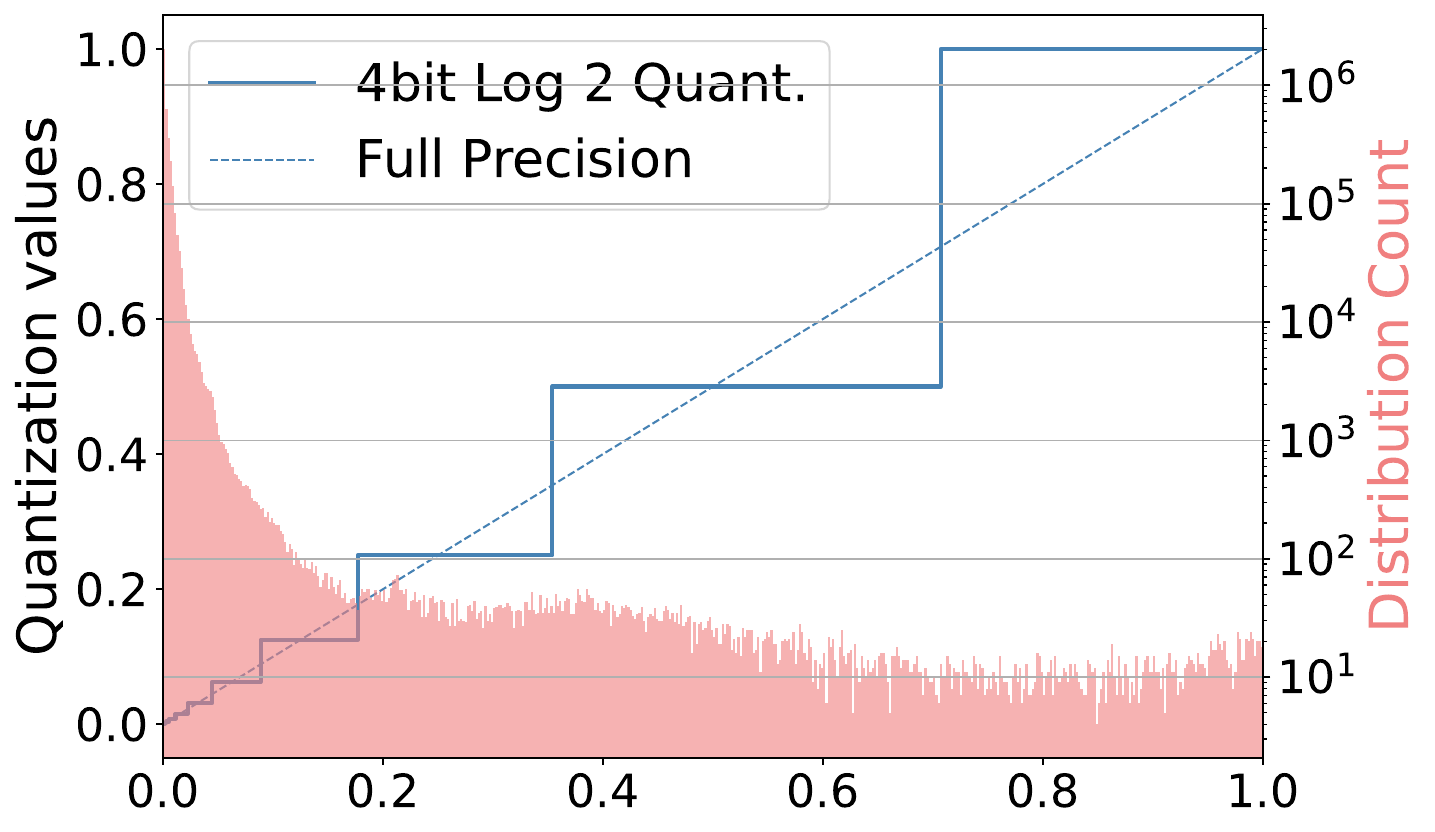}
        \caption{4-bit log2 Quantizer}
        \label{fig:sub2}
    \end{subfigure}
    \begin{subfigure}{.36\columnwidth}
        \centering
        \includegraphics[width=.98\linewidth]{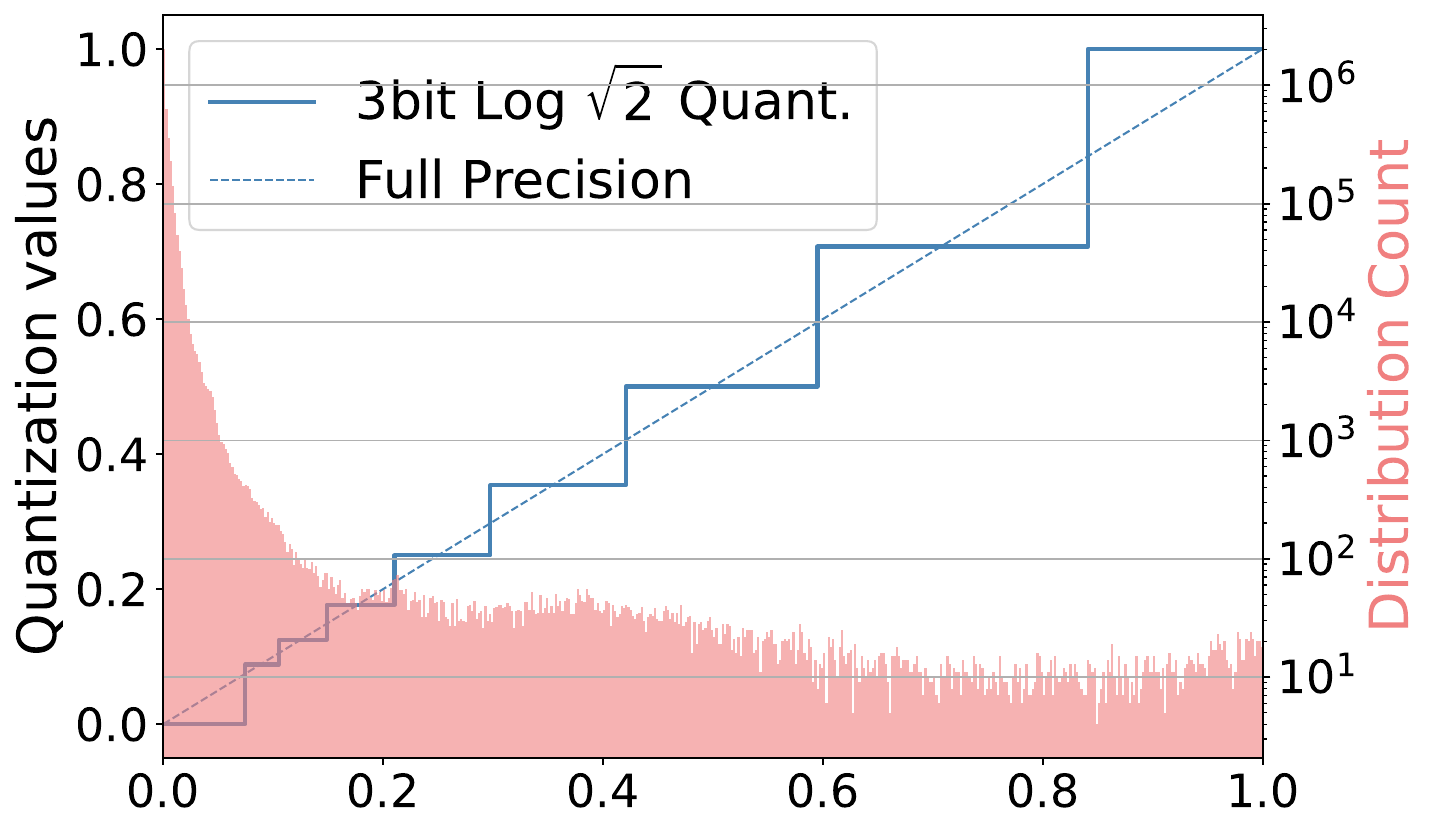}
        \caption{3-bit log$\sqrt{2}$ Quantizer}
        \label{fig:sub3}
    \end{subfigure}%
    \begin{subfigure}{.36\columnwidth}
        \centering
        \includegraphics[width=.98\linewidth]{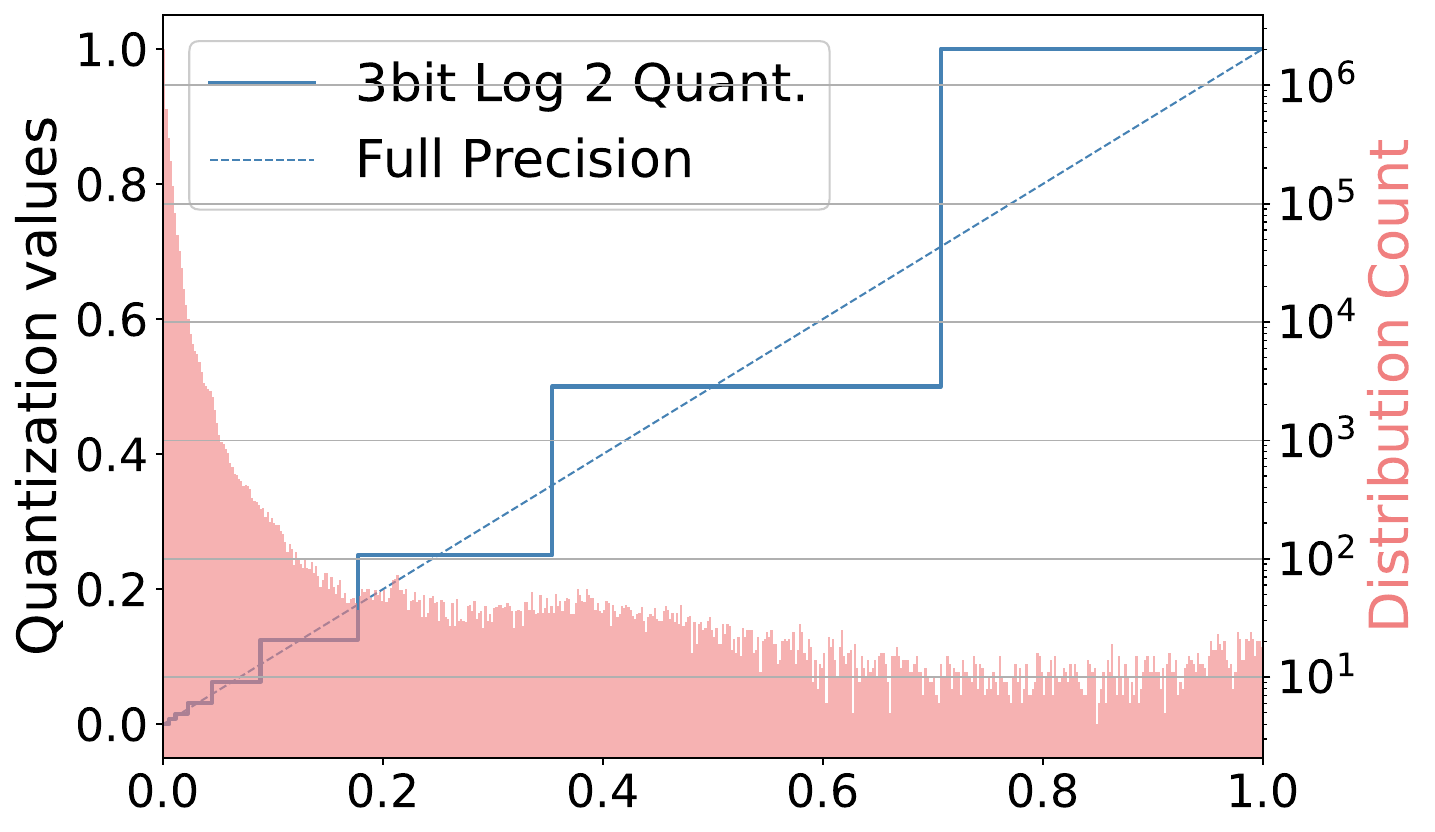}
        \caption{3-bit log2 Quantizer}
        \label{fig:sub4}
    \end{subfigure}
    \caption{Histogram of post-Softmax activations. (a)-(b): In 4-bit quantization, the log$\sqrt{2}$ quantizer allocates more bits to the relatively important large values compared to the log2 quantizer, thus reaching higher accuracy. (c)-(d): In 3-bit quantization, the log$\sqrt{2}$ quantizer quantizes the majority of values to 0, leading to significant degradation.}
    \label{fig:log_quantizer}
\end{figure}

Along with the success of Transformers in natural language processing \cite{attention}, Vision Transformer (ViT) has become a prevailing deep neural network architecture in the computer vision community, achieving promising performance for a variety of vision tasks such as image classification \cite{vit, swin, crossvit}, object detection \cite{objdettrans, detr, dydetr, octr, catdet}, semantic segmentation \cite{segmenter, segsts, effiseg}, and action recognition \cite{cvpt}. Nonetheless, the advancement of vision transformers in accuracy comes at the cost of increased model size and slow inference speed, substantially hindering its applicability in practice, especially when deploying on resource-constrained mobile or edge devices \cite{whitepaper}. 

Recently, model quantization has emerged as an effective way to compress and accelerate deep models by mapping their weights or activations with full precision into integers of lower bit-width. Existing approaches for model quantization mainly fall into the following two categories: the Quantization Aware Training (QAT) \cite{pact, learnstep} and the Post-Train Quantization (PTQ) \cite{quantsurvey, quantsurvey2}. Despite the advantage in accuracy, QAT usually needs to retrain the model on the entire training dataset, taking a considerable amount of time and computational cost. By contrast, PTQ merely requires a small-scale validation set to obtain a quantized model with even comparable accuracy, thus being much more efficient. 

In this paper, we mainly focus on the PTQ-based approaches. Most of these methods achieve promising performance with sufficiently large bit-width, but their accuracy drops sharply when quantizing with extremely low bit-width (\eg 4 bits and below). Actually, they suffer from the following two limitations. 1) \emph{Inflexible Logarithm Base.} The representative logarithm-based non-uniform quantizers \cite{FQViT, RepQViT} adopt a fixed base, \ie 2 or $\sqrt{2}$, to deal with the power-law-like activation distributions. As shown in \cref{fig:log_quantizer}, the \(\log2\) quantizer incurs substantial rounding errors for large activations under 4-bits, while the \(\log\sqrt{2}\) quantizer suffers from truncation errors for small activations under 3-bits. Moreover, the value range of post-GELU activations differs significantly in distinct layers as displayed in \cref{fig:gelu_quantizer}. This implies that the current logarithm quantizers with fixed bases cannot adaptively search for an optimal partitioning on the truncation interval as the data or bit-width varies, thus deteriorating the ultimate accuracy. In the meantime, the \(\log\sqrt{2}\) quantizer fails to avoid the floating-point multiplication as shown in \cref{fig:adalog_dialog}(b), making it not hardware-friendly. 2) \emph{Excessively sparse partition of hyperparameter search space.} Given the wide distribution range of ViT activations, the potential value range for the corresponding scaling factor also becomes broad. The conventional grid search usually adopts a uniform sparse partitioning of the entire search space by considering the search efficiency, which however is prone to fall into a local optimum.

\begin{figure}[t]
    \centering
    \includegraphics[width=\linewidth]{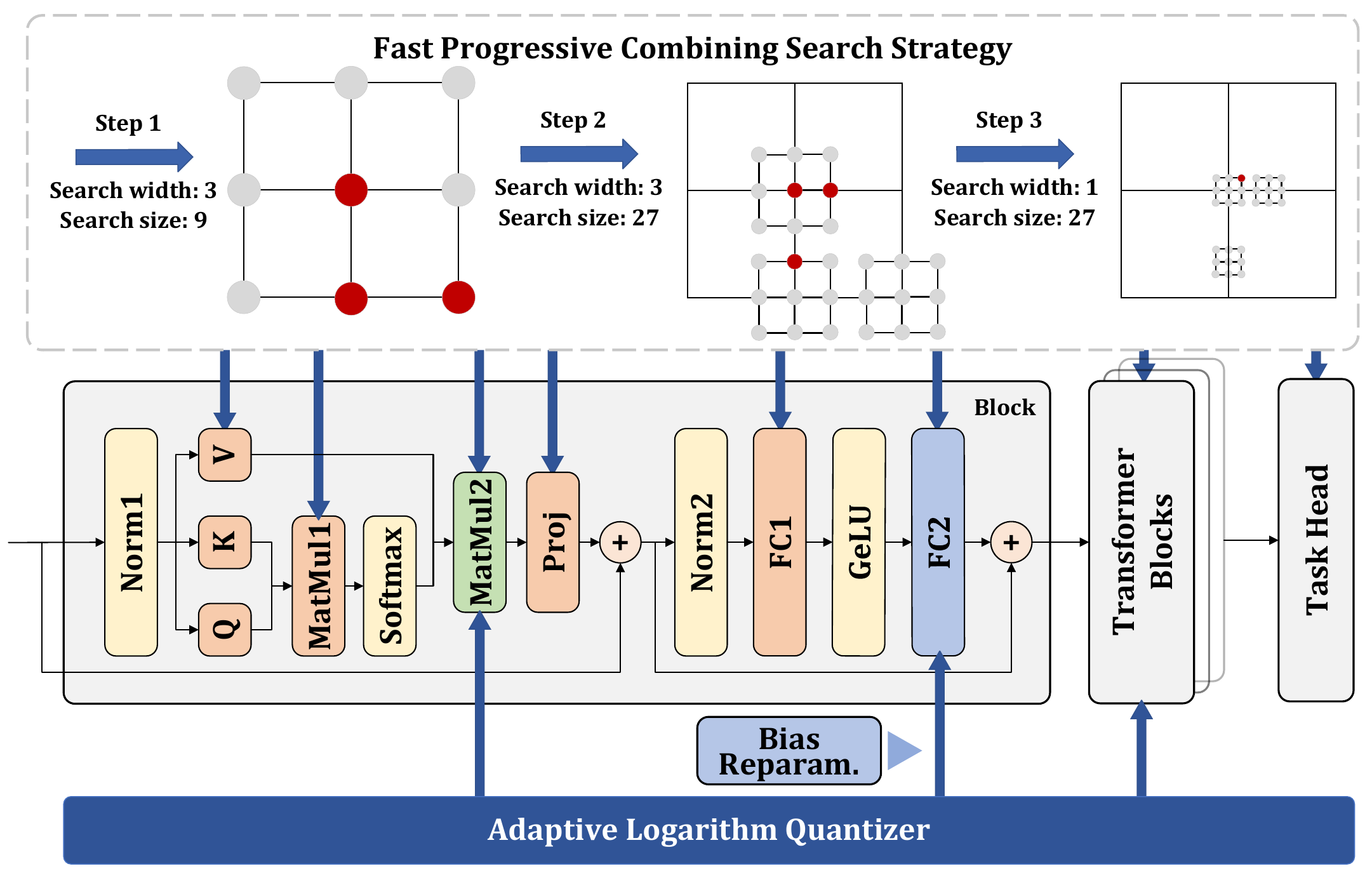}
    \caption{Illustration on the framework of our method. The AdaLog quantizer is employed for quantizing the post-Softmax and post-GELU activations, where the bias reparameterization is specifically integrated to extend AdaLog to the post-GELU layers. The Fast Progressive Combining Search (FPCS) strategy facilitates AdaLog to search for the optimal scaling factors and logarithm base, as well as the scaling factors and zero points of the uniform quantizers.}
    \label{fig:framework}
\end{figure}

To address the above issues, as illustrated in \cref{fig:framework}, we propose a novel quantizer, namely \textbf{Ada}ptive \textbf{Log}arithm (\textbf{AdaLog}) Quantizer, for post-training quantization of vision transformers. To deal with the \emph{inflexible logarithm base}, AdaLog firstly establishes the quantization and the de-quantization process with an arbitrary base, which allows for efficient computation in the integer form, thus being hardware-friendly. The optimal logarithm base is subsequently determined via hyperparameter searching. By further employing the bias reparameterization, AdaLog is applicable to quantize both the post-Softmax and post-GELU activations. To tackle the \emph{sparse hyperparameter search space}, we develop a \textbf{F}ast \textbf{P}rogressive \textbf{C}ombining \textbf{S}earch (\textbf{FPCS}) strategy that divides the search space more finely without increasing the searching complexity, compared to the previous grid search methods. Besides, this strategy can be used not only for general uniform quantizers but also for the base search of AdaLog quantizers.

The main contributions of our work are summarized in three-fold:
\begin{enumerate}
    \item We propose a novel quantizer, dubbed AdaLog, for post-training quantization of vision transformers. This non-uniform quantizer adapts the logarithmic base to accommodate the power-law distribution of activations and simultaneously allows for hardware-friendly quantization and de-quantization. It is applicable to both the post-Softmax and post-GELU activations, by employing the bias reparameterization.
    \item We develop an efficient hyperparameter search strategy, namely the Fast Progressive Combining Search strategy. Compared to the conventional uniform grid search, it is able to locate 
    the optimal hyperparameter more precisely without sacrificing the efficiency, thus being more suitable for quantizers with multiple hyperparameters.
    \item We extensively evaluate the performance of the proposed method on various computer vision tasks including classification, object detection, and instance segmentation. The experimental results demonstrate that our method significantly outperforms the state-of-the-art approaches with distinct vision transformer architectures, especially in the case of low-bit quantization. 
\end{enumerate}

\section{Related Work}
\label{sec:related}

\subsection{Vision Transformer}

Vision Transformer (ViT) and its variants have emerged as pivotal backbone networks in the computer vision community. ViT \cite{vit} firstly introduces the self-attention mechanism to the image classification task, eschewing convolutional layers in favor of Transformer blocks, thereby unveiling the potential of Transformers for visual representation learning. To enhance the efficiency, DeiT \cite{deit} integrates the knowledge distillation technique to optimize the training in the data-restricted scenario. Meanwhile, the Swin Transformer \cite{swin} adopts the hierarchical design and localized windowed self-attention to reduce the computational overhead while strengthening its capability of mining the long-range dependency. 

Due to the intensive matrix multiplication operations involved, the vision transformers usually take a considerable amount of time and memory cost, hindering their deployment in practical applications. To overcome these drawbacks, MobileViT \cite{mobilevit} and LeViT \cite{levit} attempt to design lightweight vision transformer structures. Alternatively, Token Merging \cite{tokenmerge} and X-pruner \cite{xpruner} endeavor to expedite the inference speed of ViTs through the pruning technique.

\subsection{Model Quantization}

Model quantization is a critical technique for model compression, aiming at mapping the floating-point weights and activations to integers or values of even lower bit-width. Most of the existing approaches can be categorized into Quantization-Aware Training (QAT) and Post-Training Quantization (PTQ). The QAT approaches \cite{learnstep, haq, qvit} usually achieve high accuracy, but are computationally intensive, as they retrain the model on the entire training dataset. 

By contrast, the PTQ method only needs to calibrate on small-scale data, thus being suitable for rapid deployment. AdaRound \cite{adaround}, BRECQ \cite{BRECQ}, and QDrop \cite{qdrop} are pioneering works on PTQ, but they focus on the convolutional neural networks. Several recent works have explored PTQ for vision transformers. FQ-ViT \cite{FQViT} designs a Power-of-Two Factor for LayerNorm quantization and a $\log 2$ quantizer for softmax quantization. Based on FQ-ViT, Evol-Q \cite{evolq} introduces small perturbations in quantization and utilizes an InfoNCE loss to enhance the performance. EasyQuant \cite{easyquant} employs an alternating optimization strategy for weights and activations to mitigate the quantization loss. By following EasyQuant, PTQ4ViT \cite{PTQ4ViT} introduces the Twin-Uniform quantizer to address power-law distributions and advances a Hessian-guided search strategy for optimization. APQ-ViT \cite{APQViT} boosts the Hessian-guided approach with Blockwise the Bottom-elimination Calibration and incorporates a scale parameter in the quantization of attention maps to preserve the Matthew effect. RepQ-ViT \cite{RepQViT}, focuses on quantizing the Post-LayerNorm layer by employing the reparameterization technique to balance the large activation quantization errors with small weight quantization inaccuracies. It further introduces the \(\log \sqrt{2}\) quantizer to promote the accuracy. However, current methods still severely suffer from the substantial quantization loss at low bit-width.

\section{The Proposed Approach}

\noindent \textbf{Overview~} As displayed in \cref{fig:framework}, there are four linear layers in a standard ViT block, including \textit{QKV}, \textit{Proj}, \textit{FC1} and \textit{FC2}, as well as two matrix multiplications denoted by \textit{MatMul1} and \textit{MatMul2}. Existing approaches \cite{RepQViT} have extensively studied quantizing the \textit{QKV} and \textit{FC1} layers. However, the post-Softmax and the post-GELU layers, including \textit{MatMul2} and \textit{FC2}, have not been properly handled yet, thus still suffering from non-negligible quantization loss. In this paper, we first introduce some preliminaries, and employ an adaptive logarithm quantizer (AdaLog) for post-Softmax and post-GELU layers, as detailed in Sec. \ref{sec:adalog_intro} and Sec. \ref{sec:adalog_gelu} respectively. Moreover, to address the issue of sparse partition of hyper-parameter search space in low-bit quantization, the Fast Progressive Combining Search (FPCS) strategy is employed in all the quantized layers, which is elaborated in Sec. \ref{sec:fpcs}.

\subsection{Preliminaries} \label{sec:preliminaries}

\noindent \textbf{Power-Law Distribution of Activations.~} As displayed in \cref{fig:log_quantizer}, the post-Softmax activations exhibit a power-law probability distribution, making it challenging for model quantization. The \( \log{2}\) quantizer \cite{FQViT} and \( \log{\sqrt{2}} \) quantizer \cite{RepQViT} have attempted to deal with the above problem by non-uniformly partitioning the truncation intervals with fixed logarithm bases, which are briefly described as below.

\noindent \textbf{Log2 Quantizer.~} The \(\log 2\) quantizer \cite{FQViT} is a common choice to deal with the power-law activation distributions, which can be formulated as:
\begin{align}
    \text{\emph{Quantization}}&: \bm{A}^{(\mathbb{Z})} = \text{clamp}\left(\left\lfloor -\log_2\frac{\bm{A}}{s}\right\rceil, 0, 2^{bit}-1\right), \\
    \text{\emph{De-quantization}}&: \widehat{\bm{A}} = s \cdot 2^{-\bm{A}^{(\mathbb{Z})}},
\end{align}
where \(\lfloor\cdot\rceil\) denotes the rounding function, \(s\in \mathbb{R}^+\) is the scaling factor and \textit{bit} is the quantization bit-width. By leveraging the efficient bit-shift operation, the \(\log_2(\cdot)\) function and the power-of-2 function can be implemented even faster than integer multiplication, thus being hardware-friendly.

\noindent \textbf{Log\(\sqrt{2}\) Quantizer.~} The \(\log \sqrt{2}\) quantizer \cite{RepQViT} adopts the scale reparameterization technique, formulated as below:
\begin{align}
    \text{\emph{Quantization}}&: \bm{A}^{(\mathbb{Z})} = \text{clamp}\left(\left\lfloor-2\log_2\frac{\bm{A}}{s}\right\rceil, 0, 2^{bit}-1\right), \\
    \text{\emph{De-quantization}}&: \widehat{\bm{A}} = \widetilde{S} \cdot 2^{\lfloor -\frac{\bm{A}^{(\mathbb{Z})}}{2}\rfloor},
\end{align}
where \(\widetilde{S} = s\cdot \left(\mathds{1}[x^{(\mathbb{Z})}]\cdot (\sqrt{2}-1) + 1 \right)\) is the reparameterized scale and \(\mathds{1}[\cdot]\) is a parity indicator function.

\begin{figure}[t]
    \centering
    \includegraphics[width=\linewidth]{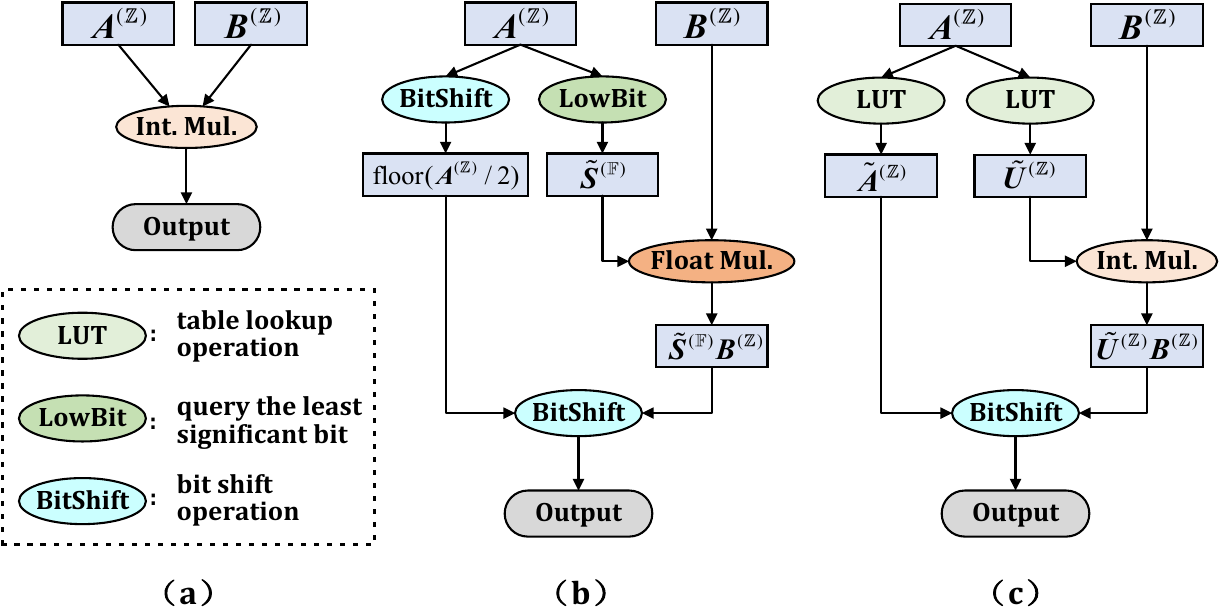}
    \caption{(a) is the flowchart for linear quantized data. (b) shows the flowchart of the $\log\sqrt{2}$ quantizer \cite{RepQViT} that fails to avoid the floating-point multiplication operation, which is not hardware-friendly. (c) displays the flowchart of the proposed AdaLog method, which only takes two extra table lookup operations and one bit-shift operation compared to the standard linear integer multiplication, making it efficient and hardware-friendly. }
    \label{fig:adalog_dialog}
\end{figure}

However, due to the differing parity of \(x^{(\mathbb{Z})}\) at various positions, \(\widetilde{S}\) becomes an element-wise floating-point scaling matrix. As shown in \cref{fig:adalog_dialog}(b), when quantizing the multiplication between two matrices $\bm{A}$ and $\bm{B}$, the \( \log{\sqrt{2}} \) quantizer needs to conduct the Hadamard product between the reparameterized scale \( \widetilde{S} \) and \( \bm{A}^{(\mathbb{Z})} \), and then multiplying with \( \bm{B}^{(\mathbb{Z})} \), where $\bm{A}^{(\mathbb{Z})}$ and $\bm{B}^{(\mathbb{Z})}$ denote the quantized form of \(\bm{A}\) and \(\bm{B}\), respectively. As it is unable to infer in the integer form, the \( \log{\sqrt{2}} \) quantizer is therefore not hardware-friendly. 

Both the \( \log{2} \) and \( \log{\sqrt{2}} \) quantizers are inflexible in searching for an optimal partition as they adopt fixed logarithm bases, thus leaving much room for improvement, especially when performing with extremely low bits (\eg 4 bits and below).

\subsection{Adaptive Logarithm Base Quantizer} \label{sec:adalog_intro}

To overcome the limitations of the log2 quantizer and the log\(\sqrt{2}\) quantizer, we propose the AdaLog quantizer by adaptively searching for an optimal logarithmic base rather than adopting a fixed one. 

Specifically, given the activation $\bm{A}$, a logarithm base $b\in \mathbb{R}^{+}$, the bit-width $bit$ and the scaling factor $s \in \mathbb{R}$, the quantization process of \(\bm{A}\) is formulated as: 
\begin{align}
\begin{split}
\textbf{\textit{A}}^{(\mathbb{Z})} &= \text{clamp}\left(\left\lfloor-\log_{b} \frac{\bm{A}}{s}\right\rceil, 0, 2^{bit}-1\right) \\
    &= \text{clamp}\left(\left\lfloor-\frac{~\log_{2}\frac{\bm{A}}{s}}{\log_2{b}} \right\rceil, 0, 2^{bit}-1\right),
\end{split}\label{eq:quant} 
\end{align}
and the de-quantization of $\textbf{\textit{A}}^{(\mathbb{Z})}$ is written as the following:
\begin{align}
\widehat{\bm{A}} = &s \cdot b^{-\bm{A}^{(\mathbb{Z})}}.\label{eq:dequant}
\end{align}

However, similar to the $\log \sqrt{2}$ quantizer, for a general base \(b\) rather than the base 2, the computation of \(b^{-\bm{A}^{(\mathbb{Z})}}\) in \cref{eq:dequant} cannot be expedited through bit shift operations. To overcome this problem, we first approximate $\log_{2}b$ using a rational number, \ie, \( \log_{2}b\approx q/r \), where $q, r \in \mathbb{Z}^+$. By applying it to \cref{eq:dequant}, the de-quantization can be reformulated as the following by employing the change-of-base formula:
\begin{equation}\label{eq:adalog}
    \widehat{\bm{A}} = s \cdot b^{-\bm{A}^{(\mathbb{Z})}} = s \cdot \left(2^{-\widetilde{\bm{A}}^{(\mathbb{Z})}} \circ 2^{-\widetilde{\bm{U}}}\right),
\end{equation}
where
\begin{align}
 \widetilde{\bm{A}}^{(\mathbb{Z})} = \left\lfloor\frac{q\cdot \bm{A}^{(\mathbb{Z})}}{r}\right\rfloor, \quad \widetilde{\bm{U}} = \frac{\left(q\cdot \bm{A}^{(\mathbb{Z})}\right)\bmod r}{r}. \label{eq:quotrem}
\end{align}

Since the elements of $\bm{A}^{(\mathbb{Z})}$ belong to $\{0,1,\cdots, 2^{bit}-1\}$, we can observe from \cref{eq:quotrem} that the elements of both $\widetilde{\bm{A}}^{(\mathbb{Z})}$ and $\widetilde{\bm{U}}$ also distribute in a finite discrete field. This implies that we can record the value range of $\widetilde{\bm{A}}^{(\mathbb{Z})}$ and $2^{-\widetilde{\bm{U}}}$ via two separate tables, once we determine the hyperparameters \(q\) and \(r\) for each layer. By this means, we only need to perform two table lookup operations to obtain \(\widetilde{\bm{A}}^{(\mathbb{Z})} \) and \( 2^{-\widetilde{\bm{U}}} \) in \cref{eq:adalog} instead of directly calculating \cref{eq:quotrem} by floating-points during the inference process.

As shown in \cref{fig:adalog_dialog}(c), the de-quantization process for the multiplication between $\bm{A}$ and $\bm{B}$ involves the operation \((2^{-\widetilde{\bm{U}}} \circ 2^{-\widetilde{\bm{A}}^{(\mathbb{Z})}})\bm{B}^{(\mathbb{Z})}\), which is not hardware-friendly as \(2^{-\widetilde{\bm{U}}}\) is a floating-point matrix. To overcome this drawback, we quantize the recorded table of \(2^{-\widetilde{\bm{U}}}\) by a uniform quantizer. Since its value falls in the range of \( (0.5,1] \) as in \cref{eq:quant_table}, we  adopt the scaling factor \( s_{\text{table}}=1/(2\cdot (2^{bit}-1)) \) and quantize $\widetilde{\bm{U}}$ into
\begin{equation} \label{eq:quant_table}
 \widetilde{\bm{U}}^{(\mathbb{Z})} = \left\lfloor \frac{2^{-\widetilde{\bm{U}}}}{s_{\text{table}}} \right\rceil.
\end{equation}

By virtue of the above steps, we can quickly obtain the reparameterized matrix \( \widetilde{\bm{U}}^{(\mathbb{Z})} \) in the integer form via the table lookup operation, and conduct the de-quantization of the multiplication between $\bm{A}$ and $\bm{B}$ in a hardware-friendly way:
\begin{equation} \label{eq:Attn_V}
\begin{split}
     \widehat{\bm{A}}\cdot\widehat{\bm{B}} = s\cdot s'\cdot s_{\text{table}} \cdot \left[\left(\widetilde{\bm{U}}^{(\mathbb{Z})} \bm{B}^{(\mathbb{Z})}\right) >> \widetilde{\bm{A}}^{(\mathbb{Z})}\right],
\end{split}
\end{equation}
where \(>>\) denotes the right shift operation, \(s\) and \(s'\) indicates the scaling factors for \(\bm{A}\) and \(\bm{B}\), respectively. The overall computational flowchart of AdaLog is 
illustrated in \cref{fig:adalog_dialog}(c).

\subsection{Extending AdaLog for Post-GELU Layers} \label{sec:adalog_gelu}

\begin{figure}[!t]
    \centering
    \begin{subfigure}{.4\columnwidth}
        \centering
        \includegraphics[width=1\linewidth]{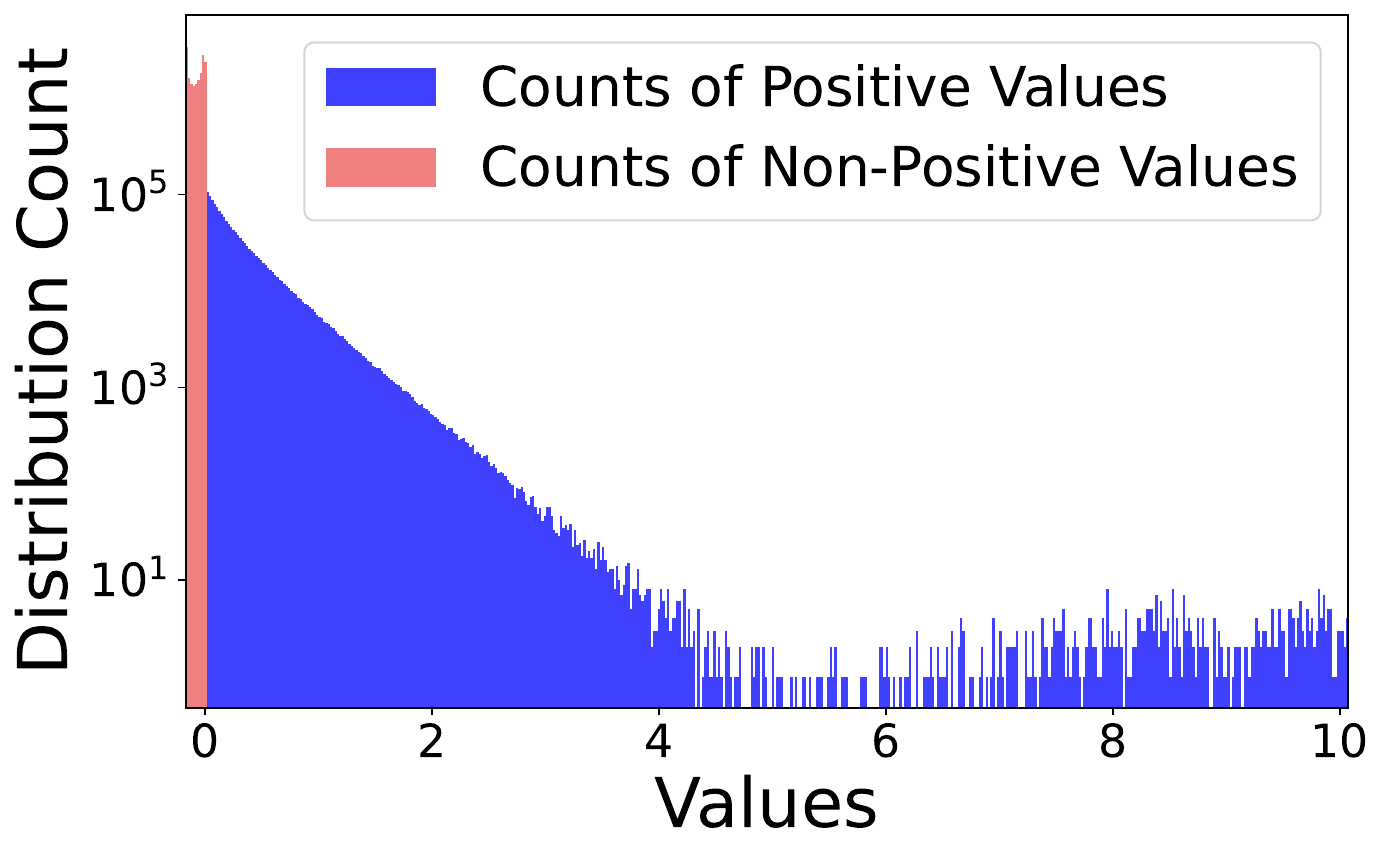}
        \caption{blocks.6.mlp.fc2}
        \label{fig:fc2_sub1}
    \end{subfigure}%
    \begin{subfigure}{.4\columnwidth}
        \centering
        \includegraphics[width=1\linewidth]{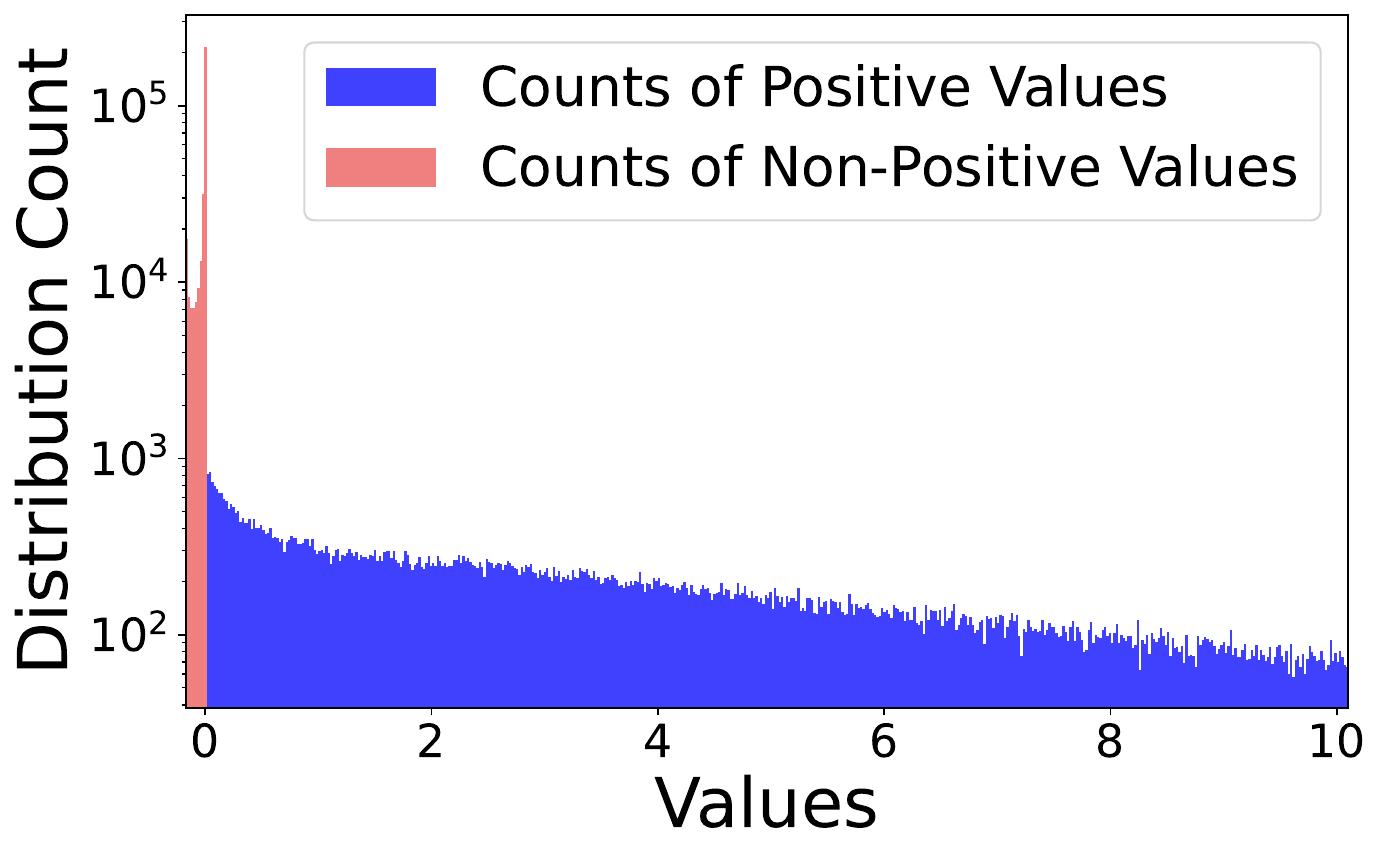}
        \caption{blocks.10.mlp.fc2}
        \label{fig:fc2_sub2}
    \end{subfigure}
    \caption{Illustration on the distribution of post-GeLU activations. (a) and (b) are the distributions of post-GeLU activation values from different layers of ViT-Base. It can be observed that although they both follow power-law-like distributions, their value ranges substantially differ, showing the necessity of adaptive logarithm bases. }
    \label{fig:gelu_quantizer}
\end{figure}

As shown in \cref{fig:gelu_quantizer}, similar to the post-Softmax layer, the activations of the post-GELU layers also obey the power-law-like distributions. Besides, the post-GELU activations suffer from the following two issues: 1) the data distributions exhibit large variations across distinct layers; 2) the majority of the values are concentrated in the range of $(-0.17, 0]$. As a consequence, AdaLog is inapplicable for the post-GELU layers as it requires non-negative inputs. To deal with the issues above, we leverage bias reparameterization to make AdaLog feasible for the post-GELU layers. 

Specifically, the post-GELU linear layer \emph{FC2} in \cref{fig:framework} has the following form:
\begin{equation}\label{eq:linear_layer}
\bm{Y} = \bm{W}\cdot\bm{X} + \bm{b},
\end{equation}
where $\bm{W} \in \mathbb{R}^{p\times m}$, $\bm{b} \in \mathbb{R}^{p}$, $\bm{X} \in \mathbb{R}^{m\times n}$ and $\bm{Y} \in \mathbb{R}^{p\times n}$ denote the weight matrix, the bias, the post-GELU activation and the output, respectively. 

Since the majority values in $\bm{X}$ locate in the range of $(-0.17, 0]$, we reformulate \cref{eq:linear_layer} as below:
\begin{equation}\label{eq:bias_reparam}
\bm{Y} = \bm{W}\cdot(\bm{X} + 0.17\cdot  \bm{1}_{m \times n}) + (\bm{b} - 0.17 \cdot \bm{W}\cdot\bm{1}_{m}),
\end{equation}
where $\bm{1}_{m}$ and $\bm{1}_{m\times n}$ refer to the $m$-dimensional vector and \(m\times n\) matrix with all ones, respectively.

As $\bm{X}'=\bm{X} + 0.17\cdot  \bm{1}_{m \times n}$ is non-negative, the proposed AdaLog is thus applicable for quantizing the first term of the linear layer in \cref{eq:bias_reparam}. Concretely, the quantization and the de-quantization are conducted as below:
\begin{align}
    {\bm{X}'}^{(\mathbb{Z})} &= \text{clamp}\left(\left\lfloor-\log_b{\frac{\bm{X}'}{s}}\right\rceil, 0, 2^{bit}-1\right), \\
    \widehat{\bm{X}}' &= s\cdot b^{-{\bm{X}'}^{(\mathbb{Z})}} \approx \bm{X}+ 0.17\cdot \bm{1}_{m \times n}. \label{eq:gelu_dequant}
\end{align}

In regards to the second term in \cref{eq:bias_reparam}, in order to keep consistent with the de-quantization on the weight $\widehat{\bm{W}}$, we employ the following bias reparameterization w.r.t. $\bm{b}$:
\begin{equation}
    \bm{b}_{rep} = \bm{b} - 0.17 \cdot \widehat{\bm{W}}\cdot\mathbf{1}_{m},
\end{equation}
where \(\widehat{\bm{W}}\) denotes the de-quantized weight and \(\bm{b}_{rep}\) denotes the reparameterized bias of \emph{FC2}.

\subsection{Fast Progressive Combining Search} \label{sec:fpcs}

Both the asymmetrically uniform quantizer and the proposed AdaLog quantizer have two types of hyperparameters. To determine these hyperparameters, existing works adopt the calibration step by using the brute-force search \cite{BRECQ,qdrop} or the alternating search \cite{PTQ4ViT,APQViT}. They require discretizing the continuous hyperparameter space through the uniform grid division. After discretization, the brute-force search traverses all possible combinations of the hyperparameters, while the alternating search iteratively fixes one hyperparameter and searches for the other. However, the complexity of brute-force search is \(O(nm)\), where \(n\) and \(m\) are the number of candidates for the two hyperparameters. By contrast, the alternating search achieves a complexity of \(O(n+m)\), but is prone to falling into a local minimum, resulting in a degradation of accuracy.

In this paper, we aim to leverage both the advantages of the above methods, by developing a hyperparameter search algorithm with a linear complexity that can finely partition the search space. Concretely, motivated by the beam search in NLP \cite{beamsearch}, we develop the Fast Progressive Combining Search (FPCS) strategy. Without loss of generality, we describe FPCS based on the asymmetrically uniform quantizer in the rest part.

\begin{algorithm}[!t]
\caption{Fast Progressive Combing Searching.}
\label{alg:overview}
\begin{algorithmic}[1]
    \Statex \textbf{Input:} Coefficients \(x,y,z_1, z_2, k, p\); a pretrained full-precision model; a set of calibration data \(\mathcal{D}_{calib}\); and the $l$-th layer to be quantized \(\phi_{l}\).
    \Statex \textbf{Output:} Quantization hyperparameters \(a^{*}, b^{*}\).
    \Statex \noindent\textcolor[RGB]{82,147,141}{\# The initialization step:}
    \State Generate the raw input $\bm{X}_{l}$ and output $\bm{O}_{l}$ by \(\phi_{l}\) based on \(\mathcal{D}_{calib}\), and compute the percentiles $pct_{0}$, $pct_{0.1}$, $pct_{0.9}$ and $pct_{1}$ by \cite{fullquant}.
    \State Compute the uniform partition of the first and second hyperparameters as $\mathcal{A}=\{ pct_{0.1}+i\cdot \tau_{A}|i=0,\cdots, x\}$ and $\mathcal{B}=\{ pct_{0.9}+j\cdot \tau_{B}|j=0,\cdots, y\}$ with the intervals $\tau_{A}=(pct_{0} - pct_{0.1}) / x$ and $ \tau_{B}=(pct_{1} - pct_{0.9}) / y$. 
     \State Generate the candidate set $\mathcal{C}_{0}$ as the Cartesian product of $\mathcal{A}$ and $\mathcal{B}$: $\mathcal{C}_{0}=\mathcal{A} \times \mathcal{B}$.
    \Statex \noindent \textcolor[RGB]{82,147,141}{ \# The progressive searching step: }
    \For{$i = 0$, $\cdots$, $p$} 
        \Statex \noindent\textcolor[RGB]{82,147,141}{\# The coarse searching step: }
        \State \parbox[t]{\dimexpr\linewidth-\algorithmicindent}{Construct the subset $\mathcal{C}'\subset \mathcal{C}_i$ by selecting the partitions that have the top-$k$ smallest quantization loss.}
        \noindent\textcolor[RGB]{82,147,141}{\# The expanding step: }
        \State Update the intervals for fine partitions:  $\tau_{A}:=\tau_{A} / (2\cdot z_1)$, $\tau_{B}:=\tau_{B} / (2\cdot z_2)$.
        \State \parbox[t]{\dimexpr\linewidth-\algorithmicindent}{Update the candidate set with fine partitions:  $\mathcal{C}_{i+1}=\{(a + i\cdot \tau_{A}, b + j\cdot \tau_{B})|(a,b)\in \mathcal{C}'; i = -z_1,\cdots,z_1; j = -z_2,\cdots,z_2\}$.}
    \EndFor
    \State The optimal hyperparameter $(a^{*}, b^{*})\in \mathcal{C}_p$ is the one that has the smallest quantization loss.
\end{algorithmic}
\end{algorithm}

\textbf{(1) Initialization.~} Assuming that the desired search complexity is \(O(n)\), we generate \(x\) candidates for the first hyperparameter and \(y\) candidates for the second one by ensuring that \(xy=n\). When quantizing the $l$-th layer $\phi_{l}$, we first utilize the calibration set $\mathcal{D}_{calib}$ to obtain the raw input $\bm{X}_{l}$ and the output $\bm{O}_{l}$ by $\phi_{l}$. We then calculate the 0.1'th and 0.9'th percentiles denoted by $pct_{0.1}$ and $pct_{0.9}$ via the Percentile 
method \cite{fullquant}. We employ a uniform partitioning scheme to derive the initial candidate set $\mathcal{C}_{0}$ for the hyperparameter search.

\textbf{(2) Coarse searching.~} Given the candidates $\mathcal{C}_{0}$, we compute the quantization loss for each candidate $(a,b)\in \mathcal{C}_{0}$, \ie the MSE loss as between the quantized output and the full-precision output denoted as $\textrm{MSE}(\phi(l)(\bm{X}_{l}, a, b), \bm{O}_{l})$, and build the subset $\mathcal{C}'\subset \mathcal{C}_{0}$ by selecting the ones with the top-\(k\) smallest losses.

\textbf{(3) Expanding.~} For each candidate $(a,b)\in \mathcal{C}'$, we expand it to \(z\) candidates with fine-grained partitions, via extending $a$ to \(z_1\) candidates and $b$ to \(z_2\) candidates, by ensuring that \(z_1z_2=z\) and \(kz=n\). The expanded candidates form the updated candidate set for searching.

\textbf{(4) Progressive searching.~} We iteratively repeat the above two steps until reaching the maximal step $p$. In the last iteration, we choose the one that has the smallest quantization loss as the optimal hyperparameter $(a^{*},b^{*})$.

The overall pipeline of the fast progressive combining search strategy is summarized in Algorithm. \ref{alg:overview}.

\section{Experimental Results and Analysis}\label{sec:experiment}

In this section, we evaluate the effectiveness of our method by comparing to the state-of-the-art post-training quantization approaches for vision transformers on the image classification task, as well as extensively conducting ablation studies and efficiency analysis. For more results on the object detection and instance segmentation tasks on COCO \cite{coco}, please refer to the \emph{Supplementary Material}.

\subsection{Experimental Setup}

\noindent \textbf{Datasets and Models.}~By following \cite{RepQViT,APQViT,PTQ4ViT}, for the classification task, we evaluate our method on ImageNet~\cite{imagenet} with representative vision transformer architectures, including ViT~\cite{vit}, DeiT~\cite{deit} and Swin~\cite{swin}. 

\noindent \textbf{Implementation Details.}~In order to make fair comparisons, we adopt the same calibration strategy as depicted in \cite{PTQ4ViT,APQViT,RepQViT}. Concretely, we randomly select 32 unlabeled images from ImageNet for the classification task. As for weights, we employ the channel-wise quantization. As for activations, we utilize layer-wise quantization in conjunction with the scale reparameterization technique. The AdaLog quantizer is used in all the post-Softmax and post-GELU activations.
We set \(r=37\) and search for the best \(q\) in \cref{eq:quotrem} by Algorithm. \ref{alg:overview}. It is worth noting that we suggest \(r\) to be a prime number such that \(q\) and \(r\) are coprime, since the value of \(\widetilde{U}\) is desired to vary when \(\bm{A}^{(\mathbb{Z})}\) takes different values. The hyperparameter \(n\) that controls the searching complexity and the searching step \(p\) in FPCS are fixed to 128 and 4, respectively. 

\subsection{Comparison with the State-of-the-Art Approaches}

We firstly compare our method with the state-of-the-art post-training quantization approaches for vision transformers on ImageNet, including PTQ4ViT~\cite{PTQ4ViT}, APQ-ViT~\cite{APQViT} and RepQ-ViT~\cite{RepQViT}. We report the results under the 6, 4, and 3 bit-widths with distinct vision transformer architectures. 

As summarized in Table~\ref{tab:classification}, for 6-bit quantization, many approaches such as PTQ4ViT and APQ-ViT exhibit a clear decrease in accuracy. In contrast, our method still delivers a promising performance, reaching the highest accuracy among the compared approaches with various backbone models. In regards to 4-bit quantization, all the compared methods suffer a remarkable degradation in accuracy due to severe quantization loss of weights and activations. However, the performance of the proposed AdaLog remains competitive in comparison with the full-precision models. Meanwhile, AdaLog significantly outperforms the compared approaches, achieving an average improvement of 5.13\% over the second best method, \ie RepQ-ViT. In addition, we evaluate on the more challenging 3-bit quantization. As displayed in Table~\ref{tab:classification}, RepQ-ViT and PTQ4VIT fail to properly deal with the quantization on the post-GELU and post-Softmax activations, thus yielding extremely low performance (\eg 0.1\%) in most scenarios. By contrast, AdaLog reaches more reasonable accuracies when reducing the bit-width from 32 to 3.

\begin{table}[!t]
    \centering
    \caption{Comparison of the top-1 accuracy (\%) on the ImageNet dataset with different quantization bit-width. `-' implies that the result is not reported or not available.}
    \begin{tabular}{ccc@{\hspace{6pt}}|@{\hspace{6pt}}c@{\hspace{6pt}}c@{\hspace{6pt}}c@{\hspace{6pt}}}
    \toprule
        Model & Full Prec. & Method & W3/A3 & W4/A4 & W6/A6 \\
        \midrule
        \multirow{4}{*}{\begin{minipage}{2cm}\centering ViT-S/224 \end{minipage} } & \multirow{4}{*}{\begin{minipage}{2cm}\centering 81.39 \end{minipage} } & PTQ4ViT & 0.10 & 42.57 & 78.63 \\
        ~ & ~ & APQ-ViT  & - & 47.95 & 79.10 \\
        ~ & ~ & RepQ-ViT  & 0.10 & 65.05 & 80.43 \\
        ~ & ~ & \textbf{AdaLog (Ours)} & \textbf{13.88} & \textbf{72.75} & \textbf{80.91} \\
        \midrule
        \multirow{4}{*}{\begin{minipage}{2cm}\centering ViT-B/224 \end{minipage} } & \multirow{4}{*}{\begin{minipage}{2cm}\centering 84.54 \end{minipage} } & PTQ4ViT & 0.10 & 30.69 & 81.65 \\
        ~ & ~ & APQ-ViT & - & 41.41 & 82.21 \\
        ~ & ~ & RepQ-ViT & 0.10 & 68.48 & 83.62 \\
        ~ & ~ & \textbf{AdaLog (Ours)} & \textbf{37.91} & \textbf{79.68} & \textbf{84.80} \\
        \midrule
        \multirow{4}{*}{\begin{minipage}{2cm}\centering DeiT-T/224 \end{minipage} } & \multirow{4}{*}{\begin{minipage}{2cm}\centering 72.21 \end{minipage} } & PTQ4ViT & 3.50 & 36.96 & 69.68 \\
        ~ & ~ & APQ-ViT & - & 47.94 & 70.49 \\
        ~ & ~ & RepQ-ViT & 0.10 & 57.43 & 70.76 \\
        ~ & ~ & \textbf{AdaLog (Ours)} & \textbf{31.56} & \textbf{63.52} & \textbf{71.38}
        \\
        \midrule
        \multirow{4}{*}{\begin{minipage}{2cm}\centering DeiT-S/224 \end{minipage} } & \multirow{4}{*}{\begin{minipage}{2cm}\centering 79.85 \end{minipage} } & PTQ4ViT & 0.10 & 34.08 & 76.28 \\
        ~ & ~ & APQ-ViT & - & 43.55 & 77.76 \\
        ~ & ~ & RepQ-ViT & 0.10 & 69.03 & 78.90 \\
        ~ & ~ & \textbf{AdaLog (Ours)} & \textbf{24.47} & \textbf{72.06} & \textbf{79.39} \\
        \midrule
        \multirow{4}{*}{\begin{minipage}{2cm}\centering DeiT-B/224 \end{minipage} } & \multirow{4}{*}{\begin{minipage}{2cm}\centering 81.80 \end{minipage} } & PTQ4ViT & 31.06  & 64.39 & 80.25 \\
        ~ & ~ & APQ-ViT & - & 67.48 & 80.42 \\
        ~ & ~ & RepQ-ViT & 0.10 & 75.61 & 81.27 \\
        ~ & ~ & \textbf{AdaLog (Ours)} & \textbf{57.45} & \textbf{78.03} & \textbf{81.55} \\
        \midrule
        \multirow{4}{*}{\begin{minipage}{2cm}\centering Swin-S/224 \end{minipage} } & \multirow{4}{*}{\begin{minipage}{2cm}\centering 83.23 \end{minipage} } & PTQ4ViT & 28.69 & 76.09 & 82.38 \\
        ~ & ~ & APQ-ViT & - & 77.15 & 82.67 \\
        ~ & ~ & RepQ-ViT & 0.10 & 79.45 & 82.79 \\
        ~ & ~ & \textbf{AdaLog (Ours)} & \textbf{64.41} & \textbf{80.77} & \textbf{83.19} \\
        \midrule
        \multirow{4}{*}{\begin{minipage}{2cm}\centering Swin-B/224 \end{minipage} } & \multirow{4}{*}{\begin{minipage}{2cm}\centering 85.27 \end{minipage} } & PTQ4ViT & 20.13  & 74.02 & 84.01 \\
        ~ & ~ & APQ-ViT & - & 76.48 & 84.18 \\
        ~ & ~ & RepQ-ViT & 0.10 & 78.32 & 84.57 \\
        ~ & ~ & \textbf{AdaLog (Ours)} & \textbf{69.75} & \textbf{82.47} & \textbf{85.09} \\
    \bottomrule
    \end{tabular}
    \label{tab:classification}
\end{table}

\subsection{Ablation Study}

\textbf{Effect of the Main Components.} We first evaluate the effectiveness of the proposed AdaLog quantizer and the FPCS strategy. As summarized in Table~\ref{tab:ablation_main},
by applying AdaLog to the post-GELU and post-Softmax activation quantization, the top-1 accuracy is significantly promoted for distinct vision transformer architectures and different bit-widths. For instance, AdaLog improves the accuracy by 9.81\%, and 4.86\% when quantizing ViT-S and DeiT-T on W4/A4, respectively. The proposed FPCS also consistently boosts the accuracy, obtaining 5.82\% and 5.02\% performance gains when quantizing ViT-B and Swin-B on W3/A3, respectively. A combination of them further promotes the performance.

\begin{table}[tp]
\setlength\tabcolsep{5pt}
\centering
\caption{Ablation results w.r.t the top-1 accuracy (\%) of the proposed main components on ImageNet with the W4/A4 and W3/A3 settings.}
\begin{tabular}{cc|ccccccc}
    \toprule
    \multirow{2} * {\textbf{AdaLog}} & \multirow{2} * {\textbf{FPCS}} & \multicolumn{2}{c}{\textbf{ViT-S} (81.39)} & \multicolumn{2}{c}{\textbf{DeiT-T} (72.21)} & \multicolumn{2}{c}{\textbf{Swin-S} (81.80)} \\
   \cmidrule(lr){3-4}\cmidrule(lr){5-6}\cmidrule(lr){7-8}
    ~ & ~ & W3/A3 & W4/A4 & W3/A3 & W4/A4 & W3/A3 & W4/A4 \\
    \midrule
     &  & 3.51 & 62.20 & 22.73 & 58.01 & 44.65 & 78.40 \\ 
     \checkmark & & 11.40 & 72.01 & 28.41 & 62.87 & 61.50 & 80.46 \\
     & \checkmark & 3.77 & 63.14 & 24.80 & 59.93 & 44.61 & 78.79 \\
    \checkmark & \checkmark & \textbf{13.88} & \textbf{72.75} & \textbf{31.56} & \textbf{63.52} & \textbf{64.41} & \textbf{80.77} \\
   \bottomrule
   \toprule
    \multirow{2} * {\textbf{AdaLog}} & \multirow{2} * {\textbf{FPCS}} & \multicolumn{2}{c}{\textbf{ViT-B} (84.54)} & \multicolumn{2}{c}{\textbf{DeiT-S} (79.85)} & \multicolumn{2}{c}{\textbf{Swin-B} (85.27)} \\
   \cmidrule(lr){3-4}\cmidrule(lr){5-6}\cmidrule(lr){7-8}
    ~ & ~ & W3/A3 & W4/A4 & W3/A3 & W4/A4 & W3/A3 & W4/A4 \\
    \midrule
     &  & 9.68 & 76.49 & 22.49 & 69.04 & 47.18 & 80.33 \\ 
     \checkmark & & 28.80 & 79.19 & 22.81 & 71.64 & 68.97 & 82.10 \\
     & \checkmark & 15.50 & 78.08 & 23.55 & 69.23 & 52.20 & 80.67 \\
    \checkmark & \checkmark & \textbf{37.91} & \textbf{79.68} & \textbf{24.47} & \textbf{72.06} & \textbf{69.75} & \textbf{82.47} \\
   \bottomrule
\end{tabular}
\label{tab:ablation_main}
\end{table}

\begin{table}[t]
\setlength\tabcolsep{7pt}
\centering
\caption{Comparison of FixOPs/Model size under different bits.}
\begin{tabular}{cccccc}
   \toprule
   \textbf{Model} & \textbf{Bits} & \textbf{Method} & \textbf{Prec.} & \textbf{FixOPs} & \textbf{Model Size} \\
   \midrule
   \multirow{4}{*}{\begin{minipage}{2.4cm}\centering DeiT-T \\ FixOPs: 20.1B \\ Size: 21.9MB \end{minipage}}  
      & 4/4 & RepQ-ViT & 57.43 & 0.613B & 3.4MB \\
   ~  & 4/4 & AdaLog  & \textbf{63.52} & \textbf{0.539B} & 3.4MB \\
   \cmidrule(lr){2-6}
   ~  & 3/3 & RepQ-ViT & 0.10 & 0.444B & 2.7MB \\
   ~  & 3/3 & AdaLog  & \textbf{31.56} & \textbf{0.391B} & 2.7MB \\
   \bottomrule
\end{tabular}
\label{tab:fixops}
\end{table}

\noindent \textbf{On the Efficiency and Effectiveness of AdaLog.~} To display the efficiency of AdaLog, we compare to RepQ-ViT in terms of the overall model size. Additionally, we report the number of FixOP \cite{apot}, \ie one operation between an 8-bit weight and an 8-bit activation, as the evaluation metric for the inference cost. Since AdaLog completely avoids floating-point operations by quantizing the lookup table, it is more efficient than the $\log\sqrt{2}$ quantizer which requires floating-point operations during inference. Table~\ref{tab:fixops} clearly shows that AdaLog is more efficient than RepQ-ViT with almost the same model size. It is worth noting that AdaLog utilizes four lookup operations in each layer, and the length of each table is \(2^{bit}\), thus taking negligible memory cost. For instance, in 4-bit quantization on DeiT-T with 12 layers, the lookup tables only take about 3KB memory, which is less than 0.2\% of the overall quantized model size.

To further demonstrate the effectiveness of AdaLog, we implement BRECQ \cite{BRECQ} on ViT-Small by using 1024 calibration images. The results show that when using the AdaLog quantizer, BRECQ significantly benefits from training activation parameters with LSQ \cite{learnstep}. However, without the AdaLog quantizer, training activation parameters may incur a collapse of accuracy. This indicates that the AdaLog quantizer can be integrated into existing PTQ frameworks, facilitating stabilizing the activation training process under extremely low bit-width.

\begin{table}[t!]
    \setlength\tabcolsep{6pt}
    \centering
    \caption{Quantization results on ImageNet. `Optim.' refers to using the BRECQ \cite{BRECQ} optimizing strategy. `Train Act.' indicates training the scaling factor of activations by applying LSQ \cite{learnstep}, besides the defaulted optimization on the rounding parameters in AdaRound \cite{adaround}.}
    \begin{tabular}{ccccc|cc}
    \toprule
        Model & AdaLog & Imgs & Optim. & Train Act. & W3/A3 & W4/A4 \\
        \midrule
        \multirow{6}{*}{\begin{minipage}{2cm}\centering ViT-S/224 \\ 81.39 \end{minipage} } & $\times$ & 32 & $\times$ & - & 3.77 & 63.14\\
        ~ & $\times$ & 1024 & \checkmark & $\times$ & 28.19 & 69.21 \\
        ~ & $\times$ & 1024 & \checkmark & \checkmark & 0.93 & 1.93 \\
        \cmidrule{2-7}
        ~ & \checkmark & 32 & $\times$ & - & 13.88 & 72.75 \\
        ~ & \checkmark &  1024 & \checkmark & $\times$ & 37.18 & 76.48 \\
        ~ & \checkmark &  1024 & \checkmark & \checkmark & \textbf{62.50} & \textbf{77.25}  \\
    \bottomrule
    \end{tabular}
    \vspace{-0.05in}
    \label{tab:adalog_train}
\end{table}

\noindent \textbf{On the Efficiency of FPCS.~} We further validate the efficiency of FPCS by comparing to the Alternating search strategy \cite{PTQ4ViT} and the Brute Force search strategy \cite{qdrop}. As shown in \cref{tab:fpcs_effi}, due to the progressive search space partitioning with linear complexity, FPCS reaches a high accuracy as the Brute Force search, while taking extremely less time cost as the Alternating search.

\begin{table}[t!]
    \setlength\tabcolsep{4pt}
    \centering
    \small
    \caption{Comparison of the top-1 accuracy and time consumption on a single NVIDIA RTX 4090 GPU during the hyperparameter search process in quantization.}
    \vspace{-0.08in}
    \begin{tabular}{cccccc}
    \toprule
        \textbf{Model} & \textbf{Method} & \textbf{Top-1 Acc.~(\(\%\))} & \textbf{Complexity} & \textbf{GPU Min.} \\
    \midrule
        \multirow{3}{*}{\begin{minipage}{2cm}\centering DeiT-T/224 \\ (W3A3) \end{minipage} } & Alternating \cite{PTQ4ViT} & 28.41 & \(O(n)\) & 3.3 \\
        ~ & Brute Force \cite{qdrop} & 32.04 & \(O(n^2)\) & 183 \\
        ~ & FPCS (Ours) & 31.56 & \(O(pn)\) & 4.1 \\
    \midrule
        \multirow{3}{*}{\begin{minipage}{2cm}\centering DeiT-S/224 \\ (W3A3) \end{minipage} } & Alternating \cite{PTQ4ViT} & 22.17 & \(O(n)\) & 5.7 \\
        ~ & Brute Force \cite{qdrop} & 29.38 & \(O(n^2)\) & 312 \\
        ~ & FPCS (Ours) & 28.51 & \(O(pn)\) & 6.5 \\
    \bottomrule
    \end{tabular}
    \label{tab:fpcs_effi}
\end{table}

\section{Conclusion}

In this paper, we propose a novel approach for post-training quantization of vision transformers. We first develop a non-uniform quantizer dubbed AdaLog that is capable of adaptively selecting the logarithm base, and is simultaneously hardware-friendly during inference. By employing the bias reparameterization, AdaLog is applicable to quantize both the post-Softmax and the post-GELU activations, and significantly promote the performance. Moreover, we propose a Fast Progressive Combining Search strategy to improve the successive hyperparameter searching. Extensive experimental results show the efficiency and effectiveness of our approach for distinct ViT-based architectures.

\section*{Acknowledgments}
This work was partly supported by the National Key R\&D Program of China (2021ZD0110503), the National Natural Science Foundation of China (Nos. 62202 034, 62176012, 62022011), the Beijing Natural Science Foundation (No. 4242044), the Beijing Municipal Science and Technology Project (No. Z231100010323002), the Research Program of State Key Laboratory of Virtual Reality Technology and Systems, and the Fundamental Research Funds for the Central Universities. 


%
%
\bibliographystyle{splncs04}
\bibliography{main}

\newpage
\appendix
\renewcommand\thetable{\Alph{table}}
\renewcommand\thefigure{\Alph{figure}}
\setcounter{table}{0}

\section*{Supplementary Material}

In this supplementary material, we present more ablation study results of the proposed AdaLog quantizer in \cref{sec:ablation}. Besides, we provide additional experimental results on COCO \cite{coco} dataset in \cref{sec:coco}.

\section{More Ablation Study Results} \label{sec:ablation}

\subsection{Post-Softmax Quantizers}

\begin{table}[b!]
    \centering
    \caption{Ablation results (\%) of the post-Softmax quantizers with different bit-width on ImageNet using the W4/A4 setting.}
    \begin{tabular}{ccc@{\hspace{6pt}}|@{\hspace{6pt}}c@{\hspace{6pt}}c@{\hspace{6pt}}c@{\hspace{6pt}}}
    \toprule
        Model & W4/A4/S32 & Method & W4/A4/S4 & W4/A4/S3 & W4/A4/S2 \\
        \midrule
        \multirow{3}{*}{\begin{minipage}{2cm}\centering ViT-S/224 \end{minipage} } & \multirow{3}{*}{\begin{minipage}{2cm}\centering 72.87 \end{minipage} } & log2 & 56.74 & 51.88 & 0.10 \\
        ~ & ~ & log\(\sqrt{2}\)  & 54.78 & 0.10 & 0.10 \\
        ~ & ~ & \textbf{AdaLog} & \textbf{72.75} & \textbf{72.39} & \textbf{70.36} \\
        \midrule
        \multirow{3}{*}{\begin{minipage}{2cm}\centering ViT-B/224 \end{minipage} } & \multirow{3}{*}{\begin{minipage}{2cm}\centering 80.13 \end{minipage} } & log2 & 78.61 & 76.44 & 0.10 \\
        ~ & ~ & log\(\sqrt{2}\) & 78.91 & 0.10 & 0.10 \\
        ~ & ~ & \textbf{AdaLog} & \textbf{79.68} & \textbf{79.60} & \textbf{78.38} \\
        \midrule
        \multirow{3}{*}{\begin{minipage}{2cm}\centering DeiT-T/224 \end{minipage} } & \multirow{3}{*}{\begin{minipage}{2cm}\centering 63.84 \end{minipage} } & log2 & 62.91 & 60.79 & 0.10 \\
        ~ & ~ & log\(\sqrt{2}\) & 62.46 & 0.10 & 0.10 \\
        ~ & ~ & \textbf{AdaLog} & \textbf{63.52} & \textbf{62.86} & \textbf{59.92}
        \\
        \midrule
        \multirow{3}{*}{\begin{minipage}{2cm}\centering DeiT-S/224 \end{minipage} } & \multirow{3}{*}{\begin{minipage}{2cm}\centering 72.18 \end{minipage} } & log2 & 71.83 & 70.91 & 0.10 \\
        ~ & ~ & log\(\sqrt{2}\) & 71.64 & 0.10 & 0.10 \\
        ~ & ~ & \textbf{AdaLog} & \textbf{72.06} & \textbf{71.35} & \textbf{69.39} \\
        \midrule
        \multirow{3}{*}{\begin{minipage}{2cm}\centering DeiT-B/224 \end{minipage} } & \multirow{3}{*}{\begin{minipage}{2cm}\centering 78.29 \end{minipage} } & log2 & 77.82  & 77.03 & 0.10 \\
        ~ & ~ & log\(\sqrt{2}\) & 77.93 & 0.10 & 0.10 \\
        ~ & ~ & \textbf{AdaLog} & \textbf{78.03} & \textbf{77.86} & \textbf{76.50} \\
        \midrule
        \multirow{3}{*}{\begin{minipage}{2cm}\centering Swin-S/224 \end{minipage} } & \multirow{3}{*}{\begin{minipage}{2cm}\centering 81.01 \end{minipage} } & log2 & 80.81 & 80.62 & 0.10 \\
        ~ & ~ & log\(\sqrt{2}\) & \textbf{80.77} & 30.46 & 0.10 \\
        ~ & ~ & \textbf{AdaLog} & \textbf{80.77} & \textbf{80.83} & \textbf{80.62} \\
        \midrule
        \multirow{3}{*}{\begin{minipage}{2cm}\centering Swin-B/224 \end{minipage} } & \multirow{3}{*}{\begin{minipage}{2cm}\centering 82.55 \end{minipage} } & log2 & 81.87  & 81.56 & 0.10 \\
        ~ & ~ & log\(\sqrt{2}\) & 81.97 & 44.41 & 0.10 \\
        ~ & ~ & \textbf{AdaLog} & \textbf{82.47} & \textbf{82.08} & \textbf{81.63} \\
    \bottomrule
    \end{tabular}
    \label{tab:ablation_softmax}
\end{table}

To validate the effectiveness of the proposed AdaLog quantizer, we evaluate it on post-Softmax quantization in Table~\ref{tab:ablation_softmax}, comparing to the Log2 and Log$\sqrt{2}$ quantizers. We fix the bit-width of all other quantizers to W4A4, and compare the performance of different post-Softmax quantizers under various quantization bit-widths. Under the 4-bit setting, AdaLog achieves the best results in most cases, which are comparable to the full-precision ones. Under the 3-bit and 2-bit settings, the Log2 and Log$\sqrt{2}$ quantizers are prone to collapse with extremely low accuracy. In contrast, AdaLog performs much more steadily. 

\subsection{Post-GELU Quantizers}

Similarly, we compare with the alternative quantizers including Uniform \cite{RepQViT}, Twin Uniform \cite{PTQ4ViT}, Log$2$ \cite{FQViT} and Log$\sqrt{2}$ Quantizer \cite{RepQViT} on post-GELU quantization. As displayed in Table~\ref{tab:ablation_gelu}, the compared approaches exhibit fluctuating performance for distinct architectures. For instance, the Log$\sqrt{2}$ quantizer reaches the second best results with the ViT-S, Deit-T, Deit-S, and DeiT-B backbones, but degrades when quantizing ViT-B, Swin-S and Swin-B. In contrast, AdaLog steadily reaches the highest top-1 accuracy. 

\begin{table}[t]
\setlength\tabcolsep{2.3pt}
\centering
\caption{Ablation results (\%) on the post-GELU quantizers on ImageNet with the W4/A4 setting. ``T-Uniform'' is the abbreviation for the Twin-Uniform Quantizer in PTQ4ViT \cite{PTQ4ViT}. ``Rep.''  is the abbreviation for the Bias Reparametrization. The best results are highlighted in bold.}
\begin{tabular}{ccccccccc}
   \toprule
   \textbf{Method} & \textbf{Rep.} & \textbf{ViT-S} & \textbf{ViT-B} & \textbf{DeiT-T} & \textbf{DeiT-S} & \textbf{DeiT-B} & \textbf{Swin-S} & \textbf{Swin-B} \\
   \midrule
   Full-Precision & - & 81.39 & 84.54 & 72.21 & 79.85 & 81.80 & 83.23 & 85.27\\
   \midrule
   Uniform \cite{RepQViT} & $\times$ & 63.14 & 78.08 & 59.93 & 69.23 & 76.02 & 78.79 & 80.67 \\
   T-Uniform \cite{PTQ4ViT} & $\times$ & 65.29 & \underline{78.76} & 60.96 & 69.78 & 76.69 & \underline{80.51} & \underline{80.93} \\
   Log$2$ \cite{FQViT} & \checkmark & 39.83 & 71.27 & 59.33 & 66.30 & 68.53 & 80.36 & 78.95 \\
   Log$\sqrt{2}$ \cite{RepQViT} & \checkmark & \underline{72.44} & 46.16 & \underline{62.91} & \underline{70.60} & \underline{77.15} & 75.91 & 24.50 \\
\rowcolor{lightgray!45}\textbf{AdaLog} & \checkmark & \textbf{72.75} & \textbf{79.68} & \textbf{63.52} & \textbf{72.06} & \textbf{78.03} & \textbf{80.77}& \textbf{82.47} \\
   \bottomrule
\end{tabular}
\label{tab:ablation_gelu}
\end{table}

\begin{table}[t]
\centering
\caption{Comparison results on COCO for the object detection and instance segmentation tasks. AP\textsuperscript{b} and AP\textsuperscript{m} indicate AP\textsuperscript{box} and AP\textsuperscript{mask}, respectively. The best results are highlighted in bold.}
\begin{tabularx}{\linewidth}{ccYYYYYYYY}
   \toprule
   \multirow{3} * {\textbf{Method}} & \multirow{3} * {\textbf{bits(W/A)}} & \multicolumn{4}{c}{\textbf{Mask R-CNN}} & \multicolumn{4}{c}{\textbf{Cascade Mask R-CNN}} \\
   \cmidrule(lr){3-6}\cmidrule(lr){7-10}
   ~ & ~  & \multicolumn{2}{c}{\textbf{Swin-T}} & \multicolumn{2}{c}{\textbf{Swin-S}} & \multicolumn{2}{c}{\textbf{Swin-T}} & \multicolumn{2}{c}{\textbf{Swin-S}} \\
   ~ & ~ & AP\textsuperscript{b} & AP\textsuperscript{m} & AP\textsuperscript{b} & AP\textsuperscript{m} & AP\textsuperscript{b} & AP\textsuperscript{m} & AP\textsuperscript{b} & AP\textsuperscript{m} \\
   \midrule
   Full-Precision & 32/32  & 46.0 & 41.6 & 48.5 & 43.3 & 50.4 & 43.7 & 51.9 & 45.0 \\
   \midrule
   PTQ4ViT \cite{PTQ4ViT} & 4/4  & 6.9 & 7.0 & 26.7 & 26.6 & 14.7 & 13.5 & 0.5 & 0.5 \\
   APQ-ViT \cite{APQViT} & 4/4  & 23.7 & 22.6 & \textbf{44.7} & 40.1 & 27.2 & 24.4 & 47.7 & 41.1 \\
   RepQ-ViT \cite{RepQViT} & 4/4  & 36.1 & 36.0 & 44.2 & 40.2 & 47.0 & 41.1 & 49.3 & 43.1 \\
\rowcolor{lightgray!45}\textbf{AdaLog (Ours)} & 4/4  & \textbf{39.1} & \textbf{37.7} & 44.3 & \textbf{41.2} & \textbf{48.2} & \textbf{42.3}& \textbf{50.6} & \textbf{44.0} \\
    \midrule
   PTQ4ViT \cite{PTQ4ViT} & 6/6  & 5.8 & 6.8 & 6.5 & 6.6 & 14.7 & 13.6 & 12.5 & 10.8 \\
   APQ-ViT \cite{APQViT} & 6/6  & \textbf{45.4} & 41.2 & 47.9 & 42.9 & 48.6 & 42.5 & 50.5 & 43.9  \\
   RepQ-ViT \cite{RepQViT} & 6/6  & 45.1 & 41.2 & 47.8 & 43.0 & 50.0 & 43.5 & 51.4 & 44.6 \\
  \rowcolor{lightgray!45} \textbf{AdaLog (Ours)} & 6/6  & \textbf{45.4} & \textbf{41.3} & \textbf{48.0} & \textbf{43.2} & \textbf{50.1} & \textbf{43.6} & \textbf{51.7} & \textbf{44.8}  \\
   \bottomrule
\end{tabularx}
\label{tab:detection}
\end{table}

\section{Experimental Results on COCO} \label{sec:coco}

We further evaluate our method on the object detection and instance segmentation tasks on the COCO dataset, by comparing to PTQ4VIT \cite{PTQ4ViT}, APQ-ViT \cite{APQViT} and RepQ-ViT \cite{RepQViT}. In order to make fair comparisons, we follow the experimental settings as depicted in \cite{RepQViT}, and report the AP$^\text{box}$ and AP$^\text{mask}$ metrics by using the Mask R-CNN and Cascade Mask R-CNN frameworks based on the Swin-T/S backbones, respectively. As shown in Table~\ref{tab:detection}, AdaLog consistently achieves the highest AP$^\text{box}$ for object detection and AP$^\text{mask}$ for instance segmentation, when performing the 6-bit quantization. Compared to the full-precision model, AdaLog incurs less than 0.6\% loss in accuracy across different frameworks. For 4-bit quantization, AdaLog promotes AP$^\text{box}$ by 3.0\%, compared to existing approaches, when quantizing Mask R-CNN with the Swin-T backbone. Similar improvements are achieved in most cases when using the Cascade Mask R-CNN framework. 

\end{document}